\documentclass{article}

\usepackage{arxiv}

\usepackage[utf8]{inputenc} % allow utf-8 input
\usepackage[T1]{fontenc}    % use 8-bit T1 fonts
\usepackage{hyperref}       % hyperlinks
\usepackage{url}            % simple URL typesetting
\usepackage{booktabs}       % professional-quality tables
\usepackage{amsfonts}       % blackboard math symbols
\usepackage{nicefrac}       % compact symbols for 1/2, etc.
\usepackage{microtype}      % microtypography
\usepackage{lipsum}         % Can be removed after putting your text content
\usepackage{graphicx}
\usepackage{natbib}
\usepackage{doi}
\usepackage{amsmath}
\usepackage{cleveref}       % smart cross-referencing
\usepackage{makecell}

\title{Bidirectional Topic Matching: \\ Quantifying Thematic Overlap Between Corpora Through Topic Modelling}

\newif\ifuniqueAffiliation
\uniqueAffiliationtrue

\ifuniqueAffiliation % Standard variant of author block
\author{ \href{https://orcid.org/0000-0001-7841-2601}{\includegraphics[scale=0.06]{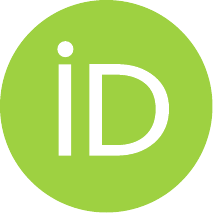}\hspace{1mm}Raven Adam} \\
	Department of Environmental Systems Sciences\\
	University of Graz\\
	Austria \\
	\texttt{} \\
	\And
	\href{https://orcid.org/0000-0002-1650-3708}{\includegraphics[scale=0.06]{orcid.pdf}\hspace{1mm}Marie Lisa Kogler}\thanks{Corresponding author} \\
	Department of Environmental Systems Sciences\
	University of Graz\\
	Austria \\
	\texttt{marie.kogler@uni-graz.at} \\
}
\else
\usepackage{authblk}

\newbox{\orcid}\sbox{\orcid}{\includegraphics[scale=0.06]{orcid.pdf}} 
\author[1]{%
	\href{https://orcid.org/0000-0001-7841-2601}{\usebox{\orcid}\hspace{1mm}Raven Adam\thanks{\texttt{hippo@cs.cranberry-lemon.edu}}}%
}
\author[1,2]{%
	\href{https://orcid.org/0000-0002-1650-3708}{\usebox{\orcid}\hspace{1mm}Marie Lisa Kogler\thanks{\texttt{marie.kogler@uni-graz.at}}}%
}
\affil[1]{Department of Computer Science, Cranberry-Lemon University, Pittsburgh, PA 15213}
\affil[2]{Department of Electrical Engineering, Mount-Sheikh University, Santa Narimana, Levand}
\fi

\renewcommand{\shorttitle}{Bidirectional Topic Matching}

\hypersetup{
pdftitle={A template for the arxiv style},
pdfsubject={cs.CL},
pdfauthor={Raven Adam, Marie Lisa Kogler},
pdfkeywords={Topic Detection, Corpus Analysis, Topic Matching},
}

\begin{document}
\maketitle

\begin{abstract}
This study introduces Bidirectional Topic Matching (BTM), a novel method for cross-corpus topic modeling that quantifies thematic overlap and divergence between corpora. BTM is a flexible framework that can incorporate various topic modeling approaches, including BERTopic, Top2Vec, and Latent Dirichlet Allocation (LDA). BTM employs a dual-model approach, training separate topic models for each corpus and applying them reciprocally to enable comprehensive cross-corpus comparisons. This methodology facilitates the identification of shared themes and unique topics, providing nuanced insights into thematic relationships.

Validation against cosine similarity-based methods demonstrates the robustness of BTM, with strong agreement metrics and distinct advantages in handling outlier topics. A case study on climate news articles showcases BTM’s utility, revealing significant thematic overlaps and distinctions between corpora focused on climate change and climate action.

BTM’s flexibility and precision make it a valuable tool for diverse applications, from political discourse analysis to interdisciplinary studies. By integrating shared and unique topic analyses, BTM offers a comprehensive framework for exploring thematic relationships, with potential extensions to multilingual and dynamic datasets. This work highlights BTM’s methodological contributions and its capacity to advance discourse analysis across various domains.
\end{abstract}

% keywords can be removed

\section{Introduction}
Topic modeling has emerged as a widely used method for analyzing and organizing large textual corpora \citep{Churchill2022}. One of its most common applications is uncovering latent topics within a given corpus, which human experts can then evaluate to derive quantitative insights \citep{Grundmann2021}. However, topic modeling is not limited to analyzing single corpora. It can also be applied to compare similarities and differences across multiple corpora \citep{Bystrov2022}. 

In the field of corpus linguistics, cross-corpus similarity methods are commonly employed, though primarily during corpus creation or selection. Nevertheless, research has demonstrated that these methods can also provide valuable insights for discourse analysis \citep{Taylor2018}. Recent studies have illustrated such applications, including comparing medical records from different hospitals \citep{Shaikina2020}, examining the economic evolution of Poland and Germany \citep{Bystrov2024}, analyzing discussions around German federal elections on Twitter \citep{Hellwig2024} and exploring migration discourse in The Times and British parliamentary debates \citep{Taylor2018}.

To advance the use of topic modeling in cross-corpus comparison, this study introduces a novel method called Bidirectional Topic Matching (BTM). BTM allows to identify shared topics across multiple corpora and understand their relationships while also highlighting corpus-specific topics. This approach facilitates both quantitative cross-corpus comparisons and the identification of potential areas for in-depth qualitative analysis.

Existing methods for cross-corpus topic modeling typically rely on topic embeddings generated by Latent Dirichlet Allocation (LDA) \citep{Blei2003} or language embedding models to compute cosine similarities between topic representations \citep{Carniel2022, Hellwig2024}. Alternatively, some studies generate a single topic model by combining corpora and then analyze differences in topic distributions across each corpus \citep{Wang2023}. In contrast, BTM employs a distinct topic model for each corpus and applies these models reciprocally across all corpora. This allows each topic model to assign topics to all documents, enabling the analysis of topic co-occurrences to identify shared and unique topics among the corpora.

\section{Method}
\label{sec:Method}
\subsection{Topic modelling}
BTM is a flexible framework for cross-corpus analysis that can incorporate various topic modeling approaches. For assessing corpus similarity, any method capable of inferring topics for new data is suitable. However, analyzing unique or corpus-specific topics requires a method that can identify intraclass outliers—documents that do not align with any topics generated by the chosen topic modeling approach. Language embedding-based methods, such as BERTopic \citep{Grootendorst2022} or Top2Vec \citep{Angelov2020}, are particularly well-suited for this purpose as they inherently support outlier detection. Traditional approaches like Latent Dirichlet Allocation (LDA), which assign a topic to every document, can also be adapted through post-processing techniques such as HDBSCAN \citep{McInnes2017} or Local Outlier Factor \citep{Breuniq2000} to identify outliers. Given BERTopic’s state-of-the-art performance and its built-in outlier detection capabilities, this study demonstrates the application and efficacy of BTM using BERTopic as the underlying topic modeling approach.

BERTopic presents an innovative method for topic modeling, capitalizing on recent advancements in embedding models. Derived from Bidirectional Encoder Representations from Transformers (BERT) \citep{Devlin2019}, this approach involves the representation of documents as points within a high-dimensional vector space. In this space, each coordinate represents contextual information corresponding to the respective document. As a result, semantically analogous documents will be in proximity to each other. Subsequently, dimensionality reduction and clustering algorithms are employed to identify compact clusters of documents with shared thematic content. Each of these clusters can then be interpreted as individual topics that are found within the investigated collection of documents and are represented by a set of keywords that are most indicative of the underlying theme. An outlier refers to a document that cannot be assigned to any of the identified topics due to its lack of thematic similarity. This occurs when the document does not align well with any of the topics, often because it is too different or semantically distant from the other documents in the model. In BERTopic, both topics and outliers can be easily accessed and handled, where outliers are grouped together under an outlier topic, often with a special identifier like -1. As a final step, a class-based term frequency inverse document frequency measure (c-TFIDF) is applied to extract the most salient terms from each topic and create interpretable topic representations \citep{Grootendorst2022}. 

\subsection{Cross-Corpus Topic Assignment}
\begin{figure}
	\centering
	{\includegraphics[scale=1]{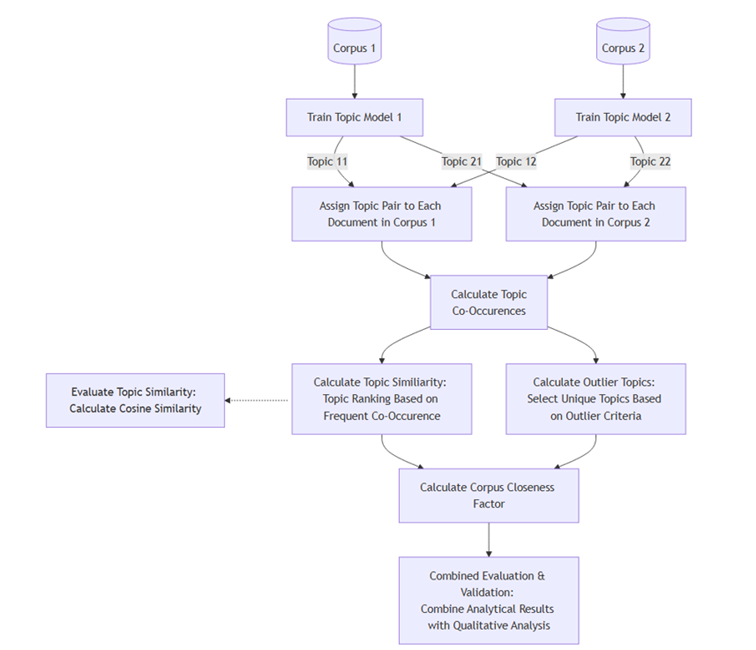}}
	\caption{Schematic Outline of Bidirectional Topic Matching Procedures for Calculating the Thematic Closeness Factor of Corpus 1 and Corpus 2. Optional additional analysis of topic similarity may be conducted via cosine similarity.}
	\label{fig:fig1}
\end{figure}

For BTM, which is schematically depictured in Figure \ref{fig:fig1}, two independent topic models are trained on two thematically related corpora, corpus 1 and corpus 2. Each model is used to identify the main themes within the respective corpus, generating topics T1 for corpus 1 and topics T2 for corpus 2. Individually, these native topic models provide a comprehensive understanding of the thematic structures specific to each dataset.

To explore thematic alignment between the corpora, each model was applied to the corpus, it was not trained on. For this, the semantic similarity between the document’s embedding and the topic embeddings of the model trained on the other corpus was calculated. Specifically, each document in corpus 2 gets matched to a topic from T1, and each document in corpus 1 gets matched to a topic from T2, based on the highest similarity score. This process produced cross-corpus topic assignments, resulting in T12 (topics from T1 assigned to Corpus 2) and T21 (topics from T2 assigned to Corpus 1). 

Subsequently, topic pairs are generated by assigning each document from one corpus to the most similar topic from the opposite corpus. Specifically, for each document, the topics assigned by the corpus 1 model (T11 and T12) and the topics assigned by the corpus 2 model (T22 and T21) are combined into cross-corpus topic pairs.

For a comprehensive cross-corpus analysis, both the main set of topics and outliers are considered. Outliers, while exhibiting atypical or low similarity scores within their own topic model, are included in the pairing process if they represent the highest similarity match for a document. Thus, topic similarity is calculated across all topics (0, 1, 2, ..., n), with outliers treated as an additional category (-1). This approach ensures that all thematical aspects are represented, even if the relationships involving outlier topics require further scrutiny in subsequent analyses. This becomes especially crucial when working with documents that are split into smaller units, like paragraphs, where certain sections may show unexpected topic associations, increasing the likelihood of outliers that require careful attention.

\subsection{Cross-Corpus Topic Pair Analysis}
The topic pairs from both corpora were analyzed through co-occurrence analysis to identify frequently paired topics between the two models. Specifically, we calculate how often each pair — composed of one topic from the corpus 1 model (T1, T12) and one from the corpus 2 model (T2, T21) — is assigned to the same document. The cross-topic co-occurrence is given by aggregation of these pairs across all documents. This process allows us to assess the frequency with which specific cross-corpus topic combinations occur together, providing insights into their thematic relationships. High-frequency pairs indicate topics from both models that were commonly associated with similar documents, reflecting thematic alignment between the corpora. 
Although the co-occurrence analysis itself remains undirected, focusing solely on the frequency of simultaneous topic occurrences within the documents, the subsequent exploration of relationships between topics from Corpus 1 and Corpus 2 is framed in a directed context. This directed approach enables a detailed investigation of the interactions and semantic linkages between the topics across the two corpora. 

\subsubsection{Interpretation of topic pairs}
The four extreme cases outlined below are provided for illustrative purposes, designed to enhance the understanding of topic co-occurrence analysis and clarify the underlying patterns. These cases are not intended to represent a final classification, but rather to offer a conceptual framework for interpreting the diverse interaction patterns between topics in the two corpora:

\textit{A native topic shows high co-occurrence with main topics from the cross-topics}.

Interpretation: This could indicate a strong thematic alignment, where the native topic is frequently associated with the central topics from the cross-topics. This suggests a robust thematic resonance between the two corpora.

\textit{A native topic shows high co-occurrence with smaller cross-topics}.

Interpretation: This may suggest that specific aspects from the native topic are often combined with more specialized, less prominent topics from the cross-topics. This could indicate a niche connection or a more nuanced thematic relationship.

\textit{A native topic shows high co-occurrence with the outlier topic from the cross-topics}.

\textbf{Interpretation:} This could indicate that the native topic appears in documents associated with more atypical or marginal topics from the cross-topics. Such connections may point to unusual thematic overlaps or aspects of other corpus that do not reflect the central themes.

\textit{The native outlier topic shows high co-occurrence with the outlier topic from the cross-topics}.

\textbf{Interpretation:} A high co-occurrence between outlier topics suggests that both corpora exhibit similar, less focused content in their nonspecific or thematically elusive documents. This connection may be incidental or superficial, lacking a clear thematic match. Rather, it reflects the conceptual ambiguity or breadth of the outlier documents in both corpora.

\textit{The native outlier topic exhibits low co-occurrence with the outlier topic from the cross-topics}. 

\textbf{Interpretation:} A low co-occurrence between outlier topics from both corpora is generally unexpected, as outlier topics typically encompass thematically broad or unspecific content that should exhibit some level of overlap across corpora. If such a case arises, it may point to several potential issues: inconsistencies in how the topic models identify and classify outlier documents, significant heterogeneity within the outlier documents themselves, or limitations in the models' ability to capture underlying semantic similarities.

\section{Topic and Corpus Measures}
For a corpus containing T native topics, a series of measures can be calculated to describe its relationship with a second corpus containing $\tilde{T}$ cross topics. A pairing strength is introduced as a quantitative measure of the degree of association between a topic from the native corpus and a topic from the cross corpus. This measure is based on the frequency of co-occurrence of the two topics within the same documents. For a topic pair $(t_i,\tilde{t}_j)$, where $t_i  | i\in\{-1,\ldots,T\}$ belongs to the native topics and $\tilde{t}_j| j\in\{-1,\ldots,\tilde{T}\}$ belongs to the cross topics, the pairing strength $S(t_i, \tilde{t}_j)$ can be defined as:
\begin{equation}
	S(t_i, \tilde{t}_j) = \frac{n(D_{ij})}{n(D_i)}
\end{equation}
where $n(D_{ij})$ denotes the size (or cardinality) of the set of documents $D_{ij}$ to which both topics $t_i$ and $\tilde{t}_j$ are assigned. Respectively, $n(D_i)$ denotes the size of the set of documents $D_i$ associated with the native topic $t_i$.

For the cross topics $\tilde{t}_j| j\in\{0,\ldots,\tilde{T}\}$, the pairing strength is referred to as topic closeness and represents the degree of alignment between each cross-topic and a specific native topic $t_i$. A special case of pairing strength involves the outlier topic $\tilde{t}_{-1}$ called topic uniqueness. Topic uniqueness quantifies the extent to which a native topic is distinct from the cross corpus. Native topics with a topic uniqueness value of 0.5 or higher are classified as unique topics.

\subsection{Corpus Closeness and Corpus Uniqueness}
Based on the topic closeness of all native topics, we define the corpus closeness $C$, which quantifies the overall thematic relatedness between the two corpora:
\begin{equation}
	C = \frac{\sum_{i=0}^{T}\sum_{j=0}^{\tilde{T}}S(t_i, \tilde{t}_j)}{T}
\end{equation}
as well as its weighted variant $C_w$, which gives higher importance to larger and more relevant native topics:
\begin{equation}
	C_w = \frac{\sum_{i=0}^{T}n(D_i)\sum_{j=0}^{\tilde{T}}S(t_i, \tilde{t}_j)}{\sum_{i=0}^{T}n(D_i)}
\end{equation}
Both closeness measures reflect the thematic overlap between the two corpora, while the weighted measure assigning greater significance to larger and thus more prominent topics within the native corpus. Generally, low closeness indicates that the two corpora are largely thematically independent.
The difference $C_w-C=\theta; \theta\in[-1,1]$ can be used to assess whether the relationship between the corpora is evenly distributed across all native topics or predominantly concentrated within a subset of native topics:
\begin{equation}
    f (x) = \left\{
    \begin{array}{ll}
    \theta\sim1; &  \text{corpus closeness is proportionally influences by larger native topics} \\
    \theta \sim 0; & \text{corpus closeness is not influenced by native topic size}\\
    \theta \sim -1 & \, \text{corpus closeness is proportionally influenced by smaller native topics} \\
    \end{array}
    \right.
\end{equation}
The corpus uniqueness $U$ and its weighted equivalent $U_w$ are alternatives to the corpus closeness to indicate the level of independence between the corpora: 
\begin{equation}
	U = 1-C=\frac{\sum_{i=0}^{T}S(t_i, \tilde{t}_{-1})}{T}
\end{equation}
\begin{equation}
	U_w = 1-C_w=\frac{\sum_{i=0}^{T}n(D_i)\cdot S(t_i, \tilde{t}_{-1})}{\sum_{i=0}^{T}n(D_i)}
\end{equation}
Here, $S(t_i, \tilde{t}_{-1})$ represents the topic uniqueness of each native topic. As with the corpus closeness factor, a high positive difference $U_w-U$ indicates that most of the corpus uniqueness is explained by larger native topics while a large negative difference sees most of it covered by smaller native topics. 

\subsection{Corpus Alignment}
Both closeness and uniqueness fail to account for the specificity of topic matches and topic size distribution of the native corpus. The topic alignment strength $SA(t_i)$ of a native topic quantifies the concentration of topic closeness values with respect to the topics of the cross corpus. This indicates whether a native topic is associated with a single theme (focused) or to multiple themes (scattered) in the other corpus. To achieve this, the highest topic closeness of the native topic is selected: 
\begin{equation}
	SA(t_i) = \max_{j\in\{0,\ldots,\tilde{T}\}}S(t_i,\tilde{t}_j)= \max\{S(t_i,\tilde{t}_0), S(t_i,\tilde{t}_1),\ldots,S(t_i,\tilde{t}_{\tilde{T}})\}
\end{equation}
A high topic alignment strength indicates that a native topic aligns with a single cross topic, whereas a low value suggests a wider variety of important pairings.

The corpus alignment $A$ serves as an overall metric that captures the average alignment strength across all native topics. It quantifies whether the topic alignments between the two corpora are focused on specific topic pairs or spread over multiple combinations.
\begin{equation}
	A = \frac{\sum_{i=0}^{T}SA(t_i)}{T}
\end{equation}
\begin{equation}
	A_w = \frac{\sum_{i=0}^{T}n(D_i)\cdot SA(t_i)}{\sum_{i=0}^{T}n(D_i)}
\end{equation}
Here, the difference $A_w-A$ is useful to indicate whether the distribution of topic alignment strength is skewed towards larger or smaller native topics.  
\subsubsection{Interpretation of corpus alignment}
By comparing the corpus uniqueness factor U and the corpus alignment factor A, we can identify three distinct edge cases describing the relationship between two corpora:

\textit{Low corpus uniqueness and low corpus alignment}.

\textbf{Interpretation:} A low uniqueness factor indicates significant thematic overlap between the corpora, while a low alignment factor suggests that most native topics are matched with multiple relevant cross topics. This implies that the cross corpus discusses the themes of the native corpus in greater detail or from multiple perspectives.

\textit{Low corpus uniqueness and high corpus alignment}.

\textbf{Interpretation:} While the corpora remain strongly related (low uniqueness), the native topics are predominantly aligned with single cross topics (high alignment). This suggests that the two corpora are likely subsets of a larger parent corpus.

\textit{High corpus uniqueness and low corpus alignment}.

\textbf{Interpretation:} A high uniqueness factor indicates that the native corpus contains many topics that are not captured by the cross corpus. Consequently, the alignment factor is low, as the matching focus values outside of topic uniqueness remain minimal. This scenario suggests that the two corpora are largely independent.

Due to the inherent relationship between the corpus uniqueness factor U and the corpus alignment factor A, a case where both values are high is not possible. 

\section{Validation through related Methods}
Since both topic models are generated from the same embedding model, the resulting embedding vectors for each topic are located in the same vector space. Therefore, to validate the effectiveness of the proposed method, we introduced an additional analysis by measuring the cosine similarity between the topic embeddings of the two BERTopic models. In this validation process, cosine similarity scores were first calculated between the topic embeddings of the corpus 1 and corpus 2 models to quantify the semantic overlap between their topics. Higher cosine similarity scores indicated greater alignment between topics. These scores were then compared to the distribution of observed topic pairs, with the goal of finding the most similar topics across the corpora. To assess the consistency between the two methods, Cohen’s kappa was calculated, providing a measure of agreement between the cosine similarity-based approach and the topic pair distribution.

\section{Case Study Climate News}
\subsection{Dataset}
To showcase BTM, two sets of digitized print articles were extracted from the WISO database that provides a repository for online newsarticles in the German-speaking region. According to Adam, Scholger, and Kogler (2023), the regional climate debate is characterized by two largely independent subject areas: climate change, which encompasses information on natural and physical impacts, dangers, and risks, and climate mitigation, which focuses on actions, socio-economic strategies, and technological solutions. The search terms climate change (“klimawandel*” where the asterisk serves as a wildcard symbol that matches any suffixes or word endings attached to the German root word "klimawandel") and climate action (“klimaschutz*”) were used to create the climate change dataset (corpus 1) and the climate action dataset (corpus 2), respectively. The investigated period spans from 2002 until 2022 and includes 21.753 articles in corpus 1 and 20.135 articles in corpus 2, with an overlap of 3.111 articles.

To account for the limited encoding length of embedding models, all articles were split into smaller parts of up to 150 words, which corresponds to the average length of German paragraphs \citep{Altpeter2015}. This was done with the help of the \textit{gsd model} available in the stanza library \citep{Qi2020}. The final dataset therefore consisted of 124.500 paragraphs.

Both BERTopic models were trained based on the \textit{German Semantic STS V2} embedding model. For corpus 1 a topic model consisting of 122 topics was generated, while corpus 2 produced a topic model with 88 topics.

\section{Results}
\subsection{Case Study}
\subsubsection{Topic Pairs and primary Relationships between Topics}
Tables \ref{tab:table1} and \ref{tab:table2} provide qualitative evidence supporting BTM’s ability to identify meaningful relationships between topics across corpora. By examining paired topics, corpus-specific nuances emerge. For example, a comparison of topics focused on forests and glaciers reveals differences in thematic emphasis: Corpus 1 highlights specific results of climate change, such as increased bark beetle infestations and rockfalls in the Alps, while Corpus 2 emphasizes the state of forests or national parks and the impact of climate change on alpine temperatures.
This capacity to reveal varying degrees of specificity allows researchers to understand how distinct datasets prioritize or converge on shared themes. Such insights are critical for comparative discourse analyses, such as political communication or cross-cultural studies.

\begin{table}
	\caption{Five native topics of corpus 1 along with their respective main cross topic pair from corpus 2 and topic alignment strength $SA$ (highest pairing strength). Each topic is represented by five topic words or phrases (connected with an underscore), which is the standard output of BERTopic. The topic representations were translated from German to English.}
	\centering
	\begin{tabular}{lll}
		\toprule
		Native Topics Corpus 1 (T1) & Cross Topics Corpus 2 (T12)& $SA$ \\
		\midrule
		EU ÖVP Austria Government & Greens ÖVP FPÖ Sebastian\_Kurz  & $0.44$     \\
		Trees Bark\_Beetle Federal\_Forestry Spruce     & Woods Hectare Federal\_Forestry National\_Park & $0.60$      \\
		Fridays Greta\_Thunberg Streets Youths     & Fridays Greta\_Thunberg Movement Humans       & $0.69$  \\
        Glacier Alps Rockfall Dachstein     & Degree Glacier Temperatures Climate\_Change       & $0.58$  \\
        Diesel Electric\_Cars Vehicles Automobile\_Industry     & Electric\_Cars Vehicles BMV Diesel       & $0.41$  \\
		\bottomrule
	\end{tabular}
	\label{tab:table1}
\end{table}

\begin{table}
	\caption{Five native topics of corpus 2 along with their respective main cross topic pair from corpus 1 and topic alignment strength $SA$ (highest pairing strength). Each topic is represented by five topic words or phrases (connected with an underscore), which is the standard output of BERTopic. The topic representations were translated from German to English.}
	\centering
	\begin{tabular}{lll}
		\toprule
		Native Topics Corpus 2 (T2) & Cross Topics Corpus 1 (T21)& $SA$ \\
		\midrule
		Greens ÖVP FPÖ Sebastian\_Kurz & EU ÖVP Austria Government  & $0.70$     \\
		Brussels Parliament Head\_of\_Government Barroso     & EU ÖVP Austria Government & $0.61$      \\
		\makecell[l]{Renovation Residential\_Construction Housing\_Subsidies \\Buildings}     & \makecell[l]{Passive\_House Residential\_Construction \\ Energy\_Efficiency Real\_Estate}      & $0.31$  \\
        ÖBB Million\_Euro Truck Commuter     & ÖBB Vienna Mobility Means\_of\_Transport       & $0.68$  \\
        Baerbock Merkel CSU Greens    & Laschet Baerbock Greens Coalition       & $0.74$  \\
		\bottomrule
	\end{tabular}
	\label{tab:table2}
\end{table}

\subsubsection{Subpairing Topics – Quantifying Secondary Thematic Relationships}
\begin{figure}
	\centering
	{\includegraphics[scale=1]{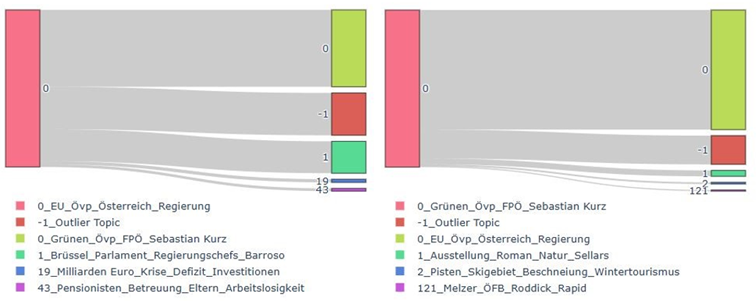}}
	\caption{Left side – The largest native topic from corpus 1 along with the five most prominent cross topic pairs from corpus 2. They gray area indicates the pairing strength for each pair. Right side – The largest native topic from corpus 2 along with the five most prominent cross topic pairs from corpus 1. They gray area indicates the pairing strength for each pair.}
	\label{fig:fig2}
\end{figure}

Whether individual topics are directly shared between corpora or whether one corpus discusses certain topics more diversely can be analyzed using topic alignment strength, as shown in Tables \ref{tab:table1} and \ref{tab:table2}. For instance, the politics topic in Corpus 1 exhibits a moderate topic alignment strength of 0.44. This indicates that several topics from Corpus 2, beyond the most similar cross-topic, address relevant aspects of this native topic. The left side of Figure \ref{fig:fig2} visually showcases this distribution across different cross-topic pairings. This suggests that political discourse is more granular in Corpus 2, allowing its topic model to recognize distinctions within documents assigned to a single topic in Corpus 1.

Conversely, Table \ref{tab:table2} reveals that both national and EU-level politics topics in Corpus 2 exhibit high topic alignment strength with the same politics topic in Corpus 1. This supports the hypothesis that political discourse in Corpus 2 is more detailed, encompassing multiple perspectives that align with a broader theme in Corpus 1.

\begin{figure}
	\centering
	{\includegraphics[scale=1]{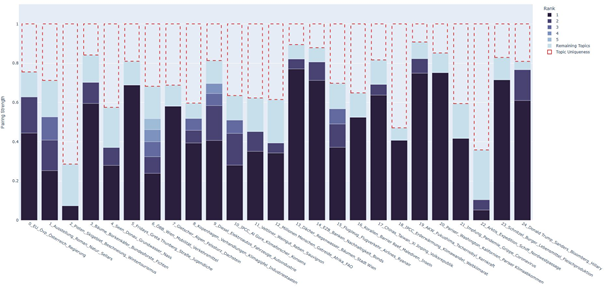}}
	\caption{The pairing strength composition for the 25 largest native topics from corpus 1. The shading of the bars indicates the ranking of the topic pairing strengths, where the most prominent pair is represented by the darkest color. Topic pairs with a pairing strength below 0.05 were merged into the “remaining topic” category. The outlier topic pairing strength or topic uniqueness is indicated by the red dashed bars.}
	\label{fig:fig3}
\end{figure}

\begin{figure}
	\centering
	{\includegraphics[scale=1]{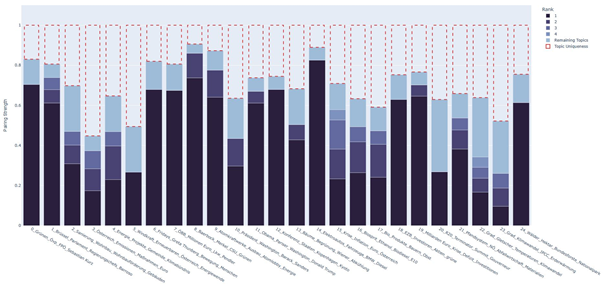}}
	\caption{The pairing strength composition for the 25 largest native topics from corpus 2. The shading of the bars indicates the ranking of the topic pairing strengths, where the most prominent pair is represented by the darkest color. Topic pairs with a pairing strength below 0.05 were merged into the “remaining topic” category. The outlier topic pairing strength or topic uniqueness is indicated by the red dashed bars.}
	\label{fig:fig4}
\end{figure}

A broader overview is provided in Figures \ref{fig:fig3} and \ref{fig:fig4}, which illustrate the pairing strength composition for the 25 largest topics in each corpus. For most native topics, the most similar cross-topic alone does not account for the majority of topic closeness. This highlights thematic asymmetries, where one corpus tends toward generality while the other emphasizes specificity. Such analyses are instrumental in uncovering where thematic overlaps or divergences occur, enabling nuanced interpretations of the data

\subsubsection{Identifying Unique Topics}
One of BTM’s most compelling features is its ability to identify topics unique to each corpus. This is achieved by extracting topics with a topic uniqueness value above 0.5. In this case study, 23 unique topics were identified in Corpus 1, while Corpus 2 contained 15 unique topics.

Table \ref{tab:table3} illustrates examples of unique topics from each corpus. Corpus 1 focuses on science communication and geographic impacts, such as water supply, while Corpus 2 emphasizes actionable measures, including renewable energy and local initiatives. Such differentiation is especially valuable for corpora with overlapping themes, as it enables researchers to discern distinct areas of focus. For example, in interdisciplinary studies, this capability bridges gaps between problem-oriented and solution-oriented approaches, fostering more comprehensive analyses.

\begin{table}
	\caption{Selection of three unique native topics from corpus 1 and corpus 2 respectively based on a topic uniqueness above 0.5.}
	\centering
	\begin{tabular}{ll}
		\toprule
		Corpus 1 Unique Topics & Corpus 2 Unique Topics \\
		\midrule
		Slopes Ski\_Area Snow\_Making Wintertourism & Austria Emissions Measures Euro\\
		Lakes Donau Groundwater Water    & Energy Project Municipality Climate\_Alliance\\
		IPCC Al\_Gore Climate\_Researcher Consensus     & Wind\_Power Renewable Austria Energie\_Transition \\
		\bottomrule
	\end{tabular}
	\label{tab:table3}
\end{table}
\subsubsection{Corpus Level Relationship}
Table \ref{tab:table4} reveals that both corpora exhibit notable distinctions, with approximately one-third of the content in each corpus not described by the other. Both show a corpus uniqueness factor of 0.34, indicating a significant level of thematic independence. The corpus closeness factor of 0.66 suggests major thematic overlaps, while the low difference between weighted and general corpus uniqueness factors $(<0.1)$ implies that neither corpus is skewed toward unique topics of particular sizes. However, Corpus 2 displays slightly more pronounced topic uniqueness in smaller topics compared to Corpus 1.

Similarly, both corpora have comparable corpus alignment factors (0.45 for Corpus 1 and 0.44 for Corpus 2). The minor influence of native topic sizes indicates that alignment is not disproportionately driven by larger topics. Together, these metrics suggest that while the corpora share substantial thematic overlap, they focus on different thematic subsets in more detail. This is consistent with the low corpus uniqueness and low corpus alignment case, where native topics frequently pair with multiple relevant cross-topics, as observed in Figures \ref{fig:fig3} and \ref{fig:fig4}.

\begin{table}
	\caption{Values for the corpus closeness factor C, the corpus uniqueness factor U, the corpus alignment factor A and the difference between the three factors and their respective weighted variants for corpus 1 and corpus 2.}
	\centering
	\begin{tabular}{lllllll}
		\toprule
		Native Corpus & $C$ & $C_w-C$ & $U$ & $u_w-U$ & $A$ & $A_w-A$ \\
		\midrule
		Corpus 1 & 0.66	& 0.02 & 0.34 & -0.02 & 0.45 & -0.01\\
		Corpus 2 & 0.66 & 0.04 & 0.34 & -0.04 & 0.44 & 0.04\\
		\bottomrule
	\end{tabular}
	\label{tab:table4}
\end{table}
\subsection{Validation - Comparison with Cosine Similarity}
We demonstrate the agreement between BTM and cosine similarity-based methods for climate news articles to highlight the validity of the proposed approach. When identifying the most similar topic from corpus 2 for each topic in corpus 1, Cohen’s kappa was calculated at 0.75. Conversely, when determining the most similar topic from corpus 1 for each topic in corpus 2, Cohen’s kappa increased to 0.81. These values reflect a strong level of agreement, affirming the reliability of BTM \citep{Mchugh2012}.

Discrepancies between BTM and cosine similarity approaches were most evident when BTM assigned the outlier topic as the closest match. Since this topic encompasses documents that do not fit into any defined clusters, its inclusion is inherently challenging for methods relying solely on cosine similarity. Beyond the outlier topic, the remaining discrepancies (approximately $20\%$) lacked clear evidence favoring one method over the other, suggesting that both approaches offer comparable utility for calculating topic similarity.
\section{Discussion and Conclusion}
BTM provides a robust framework for cross-corpus topic modeling. By leveraging BERTopic’s interpretable topic representations and employing reciprocal topic assignments, BTM facilitates a nuanced exploration of thematic relationships across corpora. This approach not only captures shared topics but also highlights unique themes, offering a comprehensive lens through which to analyze corpora with overlapping or divergent thematic structures.
\subsection{Methodological Contributions}
BTM addresses key limitations in traditional cross-corpus topic modeling approaches. By training separate topic models for each corpus and applying them reciprocally, BTM ensures that each model’s native structure is preserved while enabling cross-corpus comparisons. This dual approach allows for the identification of both shared and unique topics, a capability that is particularly valuable in interdisciplinary or comparative discourse studies.

Validation through cosine similarity underscores the reliability of BTM. Strong agreement between BTM and cosine similarity-based methods (Cohen’s kappa scores of 0.75 and 0.81) demonstrates the robustness of the approach, while the discrepancies observed with outlier topics highlight areas where BTM’s methodological strengths are most apparent. These findings suggest that BTM can serve as a reliable alternative or complement to existing methods, particularly for datasets with significant thematic variability.

\subsection{Insights from the Case Study}
The application of BTM to climate news articles revealed meaningful thematic distinctions and overlaps between two corpora focused on climate change and climate action. The results demonstrate that while both corpora share substantial thematic overlap (corpus closeness factor of 0.66), they also exhibit notable differences, with approximately one-third of the content in each corpus being unique (corpus uniqueness factor of 0.34).

Corpus 1 prioritizes broad environmental and scientific discussions, such as the geographic impacts of climate change, while Corpus 2 focuses on actionable measures like renewable energy and local initiatives. This differentiation underscores the value of BTM in identifying thematic nuances that may be overlooked by less granular methods. Moreover, the ability to quantify topic alignment and uniqueness provides a structured way to assess thematic relationships, facilitating more targeted qualitative analyses.

\subsection{Implications for Future Research}
BTM’s dual emphasis on shared and unique topics offers significant methodological advantages for various applications. In political discourse analysis, for example, BTM could reveal how different groups prioritize or frame issues, while in public health studies, it could help compare narratives across different regions or demographics. The method’s flexibility in handling diverse topic modeling approaches, including those with built-in outlier detection, further enhances its applicability across domains.

Future research could extend BTM’s utility by incorporating additional validation metrics or adapting the method for multilingual datasets. Integrating semantic similarity measures beyond cosine similarity, such as those based on contextual embeddings, may also enhance the precision of topic alignment. Additionally, applying BTM to dynamic datasets could provide insights into how thematic relationships evolve over time, offering a valuable tool for longitudinal studies.

\bibliographystyle{unsrtnat}
\bibliography{references}  %%% Uncomment this line and comment out the ``thebibliography'' section below to use the external .bib file (using bibtex) .




% 	\bibitem{keshet2016prediction}
% 	Keshet, Renato, Alina Maor, and George Kour.
% 	\newblock Prediction-Based, Prioritized Market-Share Insight Extraction.
% 	\newblock In {\em Advanced Data Mining and Applications (ADMA), 2016 12th International 
%                       Conference of}, pages 81--94,2016.

% \end{thebibliography}

\end{document}